\newif\ifFull
\Fulltrue
\ifFull
\documentclass{article}

\PassOptionsToPackage{numbers, compress}{natbib}

\usepackage{graphicx}
\usepackage{epstopdf}

\usepackage[preprint]{nips_2018}

\usepackage[utf8]{inputenc} 
\usepackage[T1]{fontenc}    
\usepackage{hyperref}       
\usepackage{url}            
\usepackage{booktabs}       
\usepackage{amsfonts}       
\usepackage{nicefrac}       
\usepackage{microtype}      
\usepackage{amsmath}

\newcommand{\E}{{\bf E}}

\newtheorem{theorem}{Theorem}

\newtheorem{definition}[theorem]{Definition}
\newenvironment{proof}{\noindent{\bf Proof:}}{\hspace*{\fill}\rule{6pt}{6pt}\bigskip}
\newcommand{\eat}[1]{}

\title{A Model for Learned Bloom Filters, \\ and Optimizing by Sandwiching}

\author{
  Michael Mitzenmacher \\
  School~of Engineering and Applied Sciences \\
  Harvard University\\
  \texttt{michaelm@eecs.harvard.edu} \\
}

\begin{document}

\maketitle
\begin{abstract}
Recent work has suggested enhancing Bloom filters by using a
pre-filter, based on applying machine learning to determine a function
that models the data set the Bloom filter is meant to represent.  Here
we model such {\em learned Bloom filters}, with the following
outcomes: (1) we clarify what guarantees can and cannot be associated
with such a structure; (2) we show how to estimate what size the
learning function must obtain in order to obtain improved performance;
(3) we provide a simple method, sandwiching, for optimizing learned Bloom
filters;  and (4) we propose a design
and analysis approach for a learned Bloomier filter, based on
our modeling approach.
\end{abstract}
\section{Introduction}
An interesting recent paper, ``The Case for Learned Index
Structures'' \cite{TCFLIS}, argues that
standard index structures and related structures, such as Bloom
filters, could be improved by using machine learning to develop what
the authors dub learned index structures.  However, this paper 
did not provide a suitable mathematical model for judging the 
performance of such structures.  Here we aim to provide a
more formal model for their variation of a Bloom filter, which they
call a {\em learned Bloom filter}.  

To describe our results, we first somewhat informally describe the learned Bloom filter. 
Like a standard Bloom filter, it provides a compressed
representation of a set of keys ${\cal K}$ that allows membership
queries.  (We may sometimes also refer to the keys as elements.)  Given a key $y$, a learned Bloom filter always
returns yes if $y$ is in ${\cal K}$, so there will be no false negatives, and generally returns
no if $y$ is not in ${\cal K}$, but may provide false positives.  
What makes a learned Bloom filter interesting is that it uses a
function that can be obtained by ``learning'' the set ${\cal K}$ to
help determine the appropriate answer;  the function acts
as a pre-filter that provides a probabilistic estimate that a query key $y$ is in ${\cal K}$.  
This learned function can be used to make an initial decision as to whether the key is
in  ${\cal K}$, and a smaller backup Bloom filter is used to prevent 
any false negatives.  

Our more formal model provides
interesting insights into learned Bloom filters, and how they might be
effective.  In particular, here we:
(1) clarify what guarantees can and cannot be associated
with such a structure; (2) show how to estimate what size the
learning function must obtain in order to obtain improved performance;
(3) provide a simple method for optimizing learned Bloom
filters;  and 
(4) demonstrate our approach may be useful for other similar structures.

We briefly summarize the outcomes above.  First, we explain how the
types of guarantees offered by learned Bloom filters differ
significantly from those of standard Bloom filters.
We thereby clarify what application-level
assumptions are required for a learned Bloom filter to be effective.
Second, we provide formulae for modeling the false positive rate for a learned Bloom
filter, allowing for an estimate of how small the learned
function needs to be in order to be effective.  We then find,
perhaps surprisingly, that a better structure uses a Bloom filter
before as well as after the learned function.
Because we optimize for two layers of Bloom filters surrounding the
learned function, we refer to this as a {\em sandwiched learned Bloom
  filter}.  We show mathematically and intuitively why sandwiching
improves performance.  We also discuss an approach to designing
learned Bloomier filters, where a Bloomier filter returns a value
associated with a set element (instead of just returning whether the
element is in the set), and show it can be modeled similarly.

While the contents of this paper may be seen as
relatively simple, we feel it is important to provide solid
foundations in order for a wide community to understand the potential
and pitfalls of data structures using machine learning components.
We therefore remark that the simplicity is purposeful, and suggest it is desirable in this context.
Finally, we note that this work incorporates and extends analysis that appeared
in two prior working notes \cite{mypaper,mypaper2}.

\section{Review:  Bloom Filters}

We start by reviewing standard Bloom filters and variants,
following the framework provided by the reference \cite{BroderMitzenmacher}.

\subsection{Definition of the Data Structure}

A Bloom filter for representing a set $S = \{x_1,x_2,\ldots,x_n\}$ of
$n$ elements corresponds to an array of $m$ bits, and uses $k$
independent hash functions $h_1,\ldots,h_k$ with range
$\{0,\ldots,m-1\}$.  Here we follow the typical assumption that these hash functions are
perfect; that is, each hash function maps each item in the universe
independently and uniformly to a number in $\{0,\ldots,m-1\}$.
Initially all array bits are 0.  For each element $x \in S$, the array
bits $h_i(x)$ are set to 1 for $1 \leq i \leq k$; it does not matter
if some bit is set to 1 multiple times.  To check if an item $y$ is in
$S$, we check whether all $h_i(y)$ are set to 1.  If not, then clearly
$y$ is not a member of $S$.  If all $h_i(y)$ are set to 1, we conclude
that $y$ is in $S$, although this may be a {\em false positive}.  A
Bloom filter does not produce false negatives.

The primary standard theoretical guarantee associated with a
Bloom filter is the following.  Let $y$ be an element
of the universe such that $y \notin S$, where $y$ is chosen
independently of the hash functions used to create the filter.  
Let $\rho$ be the fraction
of bits set to 1 after the elements are hashed.  Then
\begin{align*}
\Pr(y \mbox{ yields a false positive}) = \rho^k.
\end{align*}
For a bit in the Bloom filter to be 0, it has to not be the outcome
of the $kn$ hash values for the $n$ items.  It follows that
\begin{align*}
\E[\rho] = 1 - \left ( 1 - {1 \over m} \right )^{kn} \approx 1-{\rm e}^{-kn/m},
\end{align*}
and that via standard techniques using concentration bounds (see, e.g., \cite{MU})
\begin{align*}
\Pr(|\rho - \E[\rho]| \geq \gamma) \leq {\rm e}^{-\Theta(\gamma^2 m)}
\end{align*}
in
the typical regime where $m/n$ and $k$ are constant.  That is, $\rho$
is, with high probability, very close to its easily calculable
expectation, and thus we know (up to very small random deviations)
what the probability is that an element $y$ will be a false positive.
Because of this tight concentration around the expectation, it is
usual to talk about the {\em false positive probability} of a Bloom
filter; in particular, it is generally referred to as though it is a
constant depending on the filter parameters, even though it is a
random variable, because it is tightly concentrated around its
expectation.

Moreover, given a set of distinct query elements $Q
= \{y_1,y_2,\ldots,y_q\}$ with $Q \cap S = \emptyset$ chosen a priori
before the Bloom filter is instantiated, the fraction of false
positives over these queries will similarly be concentrated near
$\rho^k$.  Hence we may talk about the {\em false positive rate}
of a Bloom filter over queries, which (when the query elements are distinct) is essentially the
same as the false positive probability.  (When the query elements are
not distinct, the false positive rate may vary significantly,
depending on on the distribution of the number of appearances of
elements and which ones yield false positives; we focus on the distinct
item setting here.)  In particular, the false positive rate is a
priori the same for {\em any} possible query set $Q$.  Hence one
approach to finding the false positive rate of a Bloom filter
empirically is simply to test a random set of query elements (that
does not intersect $S$) and find the fraction of false positives.
Indeed, it does not matter what set $Q$ is chosen, as long as it is
chosen independently of the hash functions.

We emphasize that, as we discuss further below, the term false positive
rate often has a different meaning in the context of learning theory 
applications. Care must therefore be taken in understanding how the
term is being used.

\subsection{Additional Bloom Filter Benefits and Limitations}

For completeness, we relate some of the other benefits and limitations of
Bloom filters.  More details can be found in \cite{BroderMitzenmacher}.

We have assumed in the above analysis that the hash functions are
fully random.  As fully random hash functions are not practically
implementable, there are often questions relating to how well the
idealization above matches the real world for specific hash functions.
In practice, however, the model of fully random hash functions appears
reasonable in many cases; see \cite{Simple} for further discussion on
this point.

If an adversary has access to the hash functions used, or to the final
Bloom filter, it can find elements that lead to false positives.  One
must therefore find other structures for adversarial situations.  A
theoretical framework for such settings is developed in \cite{naor2015bloom}.
Variations of Bloom filters, which adapt to false positives
and prevent them in the future, are described in \cite{adapt2,adapt1};  while not meant
for adversarial situations, they prevent repeated false positives with the
same element.  

One of the key advantages of a standard Bloom filter is that it is
easy to insert an element (possibly slightly changing the false
positive probability), although one cannot delete an element without
using a more complex structure, such as a counting Bloom filter.  
However, there are more recent alternatives to the standard Bloom filter,
such as the cuckoo filter \cite{cuckoof}, which can achieve the same or better space 
performance as a standard Bloom filter while allowing insertions and deletions.
If the Bloom filter does not need to insert or delete elements, a well-known
alternative is to develop a perfect hash function for the data set, and store
a fingerprint of each element in each corresponding hash location (see, e.g., \cite{BroderMitzenmacher}
for further explanation);  this approach reduces the space required by approximately 30\%.  

\section{Learned Bloom Filters}

\subsection{Definition of the Data Structure}

We now consider the learned Bloom filter construction as described in
\cite{TCFLIS}.  We are given a set of positive keys ${\cal K}$ that correspond
to set to be held in the Bloom filter -- that is, ${\cal K}$ corresponds
to the set $S$ in the previous section.  We are also given a set of
negative keys ${\cal U}$ for training.  We then train a neural network with
${\cal D} = \{(x_i,y_i =1)~|~x_i \in {\cal K}\} \cup
\{(x_i,y_i =0)~|~x_i \in {\cal U}\}$;  that is, they suggest
using a neural network on this
binary classification task to produce a probability, based on minimizing
the log loss function
\begin{align*}
L = \sum_{(x,y) \in {\cal D}} y \log f(x) + (1-y)\log (1-f(x)),
\end{align*}
where $f$ is the learned model from the neural network.  Then $f(x)$
can be interpreted as a ``probability'' estimate that $x$ is a key from the set.
Their suggested approach is to choose a threshold $\tau$ so that if
$f(x) \geq \tau$ then the algorithm returns that $x$ is in the set,
and no otherwise.  Since such a process may provide false negatives
for some keys in ${\cal K}$ that have $f(x) < \tau$, a secondary
structure -- such as a smaller standard Bloom filter holding the keys 
from ${\cal K}$ that have $f(x) < \tau$ --
can be used to check keys with $f(x) < \tau$ to ensure there are no false negatives, 
matching this feature of the standard Bloom filter.

In essence, \cite{TCFLIS} suggests using a pre-filter ahead of the
Bloom filter, where the pre-filter comes from a neural network and
estimates the probability a key is in the set, allowing the use
of a smaller Bloom filter than if one just used a Bloom filter alone.  Performance improves if the size to
represent the learned function $f$ and the size of the smaller backup
filter for false negatives is smaller than the size of a corresponding
Bloom filter with the same false positive rate.  
Of course the pre-filter here need not
come from a neural network; any approach that would estimate the
probability an input key is in the set could be used.

This motivates the following formal definition:
\begin{definition}
\label{def:lbf}
A {\em learned Bloom filter} on a set of positive keys ${\cal K}$ and
negative keys ${\cal U}$ is a function $f:U \rightarrow [0,1]$ and
threshold $\tau$, where $U$ is the universe of possible query keys, and
an associated standard Bloom filter $B$, referred to as a backup filter.  The 
backup filter holds the set of keys $\{z: z \in {\cal K}, f(z) < \tau\}$.  
For a query $y$, the learned Bloom filter returns that $y \in {\cal K}$
if $f(y) \geq \tau$, or if $f(y) < \tau$ and the backup filter
returns that $y \in {\cal K}$.  The learned Bloom filter returns 
$y \notin {\cal K}$ otherwise.
\end{definition}

\subsection{Defining the False Positive Probability}
\label{sec:fpr}

The question remains how to determine or derive the false positive
probability for a learned Bloom filter, and how to choose an
appropriate threshold $\tau$.  The approach in \cite{TCFLIS} is to
find the false positive rate over a test set.  This approach is, as we
have discussed, suitable for a standard Bloom filter, where the false
positive rate is guaranteed to be close to its expected value for any
test set, with high probability.  However, this
methodology requires additional assumptions in the learned Bloom 
filter setting.

As an example, suppose the universe of elements is the range
$[0,1000000)$, and the set ${\cal K}$ of keys to store in our Bloom
filter consists of a random subset of 500 elements from the range
$[1000,2000]$, and of 500 other random elements from
outside this range.  Our learning algorithm might determine that a
suitable function $f$ yields that $f(y)$ is large (say $f(y) \approx 1/2$)
for elements in the range $[1000,2000]$ and close to zero elsewhere,
and then a suitable threshold might be $\tau = 0.4$.  The resulting
false positive rate depends substantially on what elements are
queried.  If ${\cal Q}$ consists of elements primarily in the range
$[1000,2000]$, the false positive rate will be quite high, while if
${\cal Q}$ is chosen uniformly at random over the whole range, the
false positive rate will be quite low.  Unlike a standard Bloom
filter, the false positive rate is highly dependent on the query set, and is not
well-defined independently of the queries.

Indeed, it seems plausible that in many situations, the query set
${\cal Q}$ might indeed be similar to the set of keys ${\cal K}$, so
that $f(y)$ for $y \in {\cal Q}$ might often be above naturally chosen
thresholds.  For example, in security settings, one might expect that
queries for objects under consideration (URLs, network flow features)
would be similar to the set of keys stored in the filter.  Unlike in
the setting of a standard Bloom filter, the false positive probability
for a query $y$ can depend on $y$, even before the function $f$ is instantiated.

It is worth noting, however, that the problem we point out here can
possibly be a positive feature in other settings; it might be that the
false positive rate is remarkably low if the query set is suitable.
Again, one can consider the range example above where queries are
uniform over the entire space; the query set is very unlikely to hit
the range where the learned function $f$ yields an above threshold
value in that setting for a key outside of ${\cal K}$.  The
data-dependent nature of the learned Bloom filter may allow it to
circumvent lower bounds for standard Bloom filter structures.  

While the false positive probability for learned Bloom filters does not have
the same properties as for a standard Bloom filter, 
we can define the false positive rate of a learned Bloom 
filter with respect to a given query distribution.  

\begin{definition}
\label{def:fpr}
A {\em false positive rate on a query distribution} ${\cal D}$ 
over ${\cal U} - {\cal K}$ for a learned Bloom filter $(f,\tau,B)$ is given by 
\begin{align*}
\Pr_{y \sim {\cal D}}(f(y) \geq \tau)  + (1-\Pr_{y \sim {\cal D}}(f(y) \geq \tau))F(B),
\end{align*}
where $F(B)$ is the false positive rate of the backup filter $B$.
\end{definition}
While technically $F(B)$ is itself a random variable, the false
positive rate is well concentrated around its expectations, which depends only on the size
of the filter $|B|$ and the number of false negatives from ${\cal K}$ that must be stored in
the filter, which depends on $f$.  Hence where the meaning is clear we may consider 
the false positive rate for a learned Bloom filter with function $f$ and threshold $\tau$ to be
\begin{align*}
\Pr_{y \sim {\cal D}}(f(y) \geq \tau)  + (1-\Pr_{y \sim {\cal D}}(f(y) \geq \tau))\E[F(B)],
\end{align*}
where the expectation $\E[F(B)]$ is meant to over instantiations of the Bloom filter with given size $|B|$.

Given sufficient data, we can determine an {\em empirical false
positive rate} on a test set, and use that to predict future behavior.
Under the assumption that the test set has the same distribution as
future queries, standard Chernoff bounds provide that the empirical
false positive rate will be close to the false positive rate on future
queries, as both will be concentrated around the expectation.  In many
learning theory settings, this empirical false positive rate appears to
be referred to as simply the false positive rate;  we emphasize that false
positive rate, as we have explained above, typically means something different
in the Bloom filter literature.

\begin{definition}
\label{def:efpr}
The {\em empirical false positive rate on a set} ${\cal T}$, where
${\cal T} \cap {\cal K} = \emptyset$, 
for a learned Bloom filter $(f,\tau,B)$ is the number of false positives
from ${\cal T}$ divided by $|{\cal T}|$.
\end{definition}

\begin{theorem}
\label{thmone}
Consider a learned Bloom filter $(f,\tau,B)$, a test set ${\cal T}$,
and a query set ${\cal Q}$, where ${\cal T}$ and ${\cal Q}$ are both determined
from samples according to a distribution ${\cal D}$.  
Let $X$ be the
empirical false positive rate on ${\cal T}$, and $Y$ be the
empirical false positive rate on ${\cal Q}$.  Then 
\begin{align*}
\Pr(|X - Y| \geq \epsilon) \leq {\rm e}^{-\Omega(\epsilon^2 \min(|{\cal T}|,|{\cal Q|}))}.
\end{align*}
\end{theorem}

\smallskip

\begin{proof}
Let $\alpha = \Pr_{y \sim {\cal D}}(f(y) \geq \tau)$, and $\beta$ be 
false positive rate for the backup filter.  We first 
show that for ${\cal T}$ and $X$ that
\begin{align*}
\Pr(|X - (\alpha + (1-\alpha)\beta)| \geq \epsilon) \leq 2{\rm e}^{-2\epsilon^2|{\cal T}|}.
\end{align*}
This follows from a direct Chernoff bound (e.g., \cite{MU}[Exercise 4.13]), since each sample chosen
according to ${\cal D}$ is a false positive with probability $\alpha + (1-\alpha)\beta$.  
A similar bound holds for ${\cal Q}$ and $Y$.

We can therefore conclude 
\begin{eqnarray*}
\Pr(|X - Y| \geq \epsilon) \! \! & \leq & \! \! \Pr(|X - (\alpha + (1-\alpha)\beta)| \geq \epsilon/2) \\
                           \! \! & & \! \! \mbox{ } + \Pr(|Y - (\alpha + (1-\alpha)\beta)| \geq \epsilon/2) \\
                           \! \! & \leq & \! \! 2{\rm e}^{-\epsilon^2|{\cal T}|/2} + 2{\rm e}^{-\epsilon^2|{\cal Q}|/2},
\end{eqnarray*}
giving the desired result.
\end{proof}

Theorem~\ref{thmone} also informs us that it is reasonable to find a suitable parameter $\tau$,
given $f$,  by trying a suitable finite discrete set of values for $\tau$,
and choosing the best size-accuracy tradeoff for the application.
By a union bound, all choices of $\tau$ will perform close to their expectation
with high probability.

While Theorem~\ref{thmone} requires the test set and
query set to come from the same distribution ${\cal D}$, the negative
examples ${\cal U}$ do not have to come from that distribution.  Of
course, if negative examples ${\cal U}$ are drawn from ${\cal D}$, it
may yield a better learning outcome $f$.  

If the test set and query set distribution do not match, because for
example the types of queries change after the original gathering of
test data ${\cal T}$, Theorem~\ref{thmone} offers limited guidance.
Suppose ${\cal T}$ is derived from samples from distribution ${\cal
D}$ and ${\cal Q}$ from another distribution ${\cal D'}$. If the two
distributions are close (say in $L_1$ distance), or, more
specifically, if the changes do not significantly change the
probability that a query $y$ has $f(y) \geq \tau$, then the empirical
false positive rate on the test set may still be relatively accurate.  However, in
practice it may be hard to provide such guarantees on the nature of
future queries.  This explains our previous statement that learned
Bloom filters appear most useful when the query stream can be modeled
as coming from a fixed distribution, which can be sampled during the
construction.

We can return to our previous example to understand these effects.
Recall our set consists of 500 random elements
from the range $[1000,2000]$ and 500 other random elements from the
range $[0,1000000)$.  Our learned Bloom filter has $f(y) \geq \tau$
for all $y$ in $[1000,2000]$ and $f(y) < \tau$ otherwise.  Our backup filter will
therefore store 500 elements.  If our test set is uniform over
$[0,1000000)$ (excluding elements stored in the Bloom filter), our
false positive rate from elements with too large an $f$ value would be
approximately $0.0002$; one could choose a backup filter with roughly the
same false positive probability for a total empirical false positive
probability of $0.0004$.  
If, however, our queries
are uniform over a restricted range $[0,100000)$, then the false positive
probability would jump to $0.0022$ for the learned Bloom filter, because
the learned function would yield more false positives over the smaller query
range.  

\subsection{Additional Learned Bloom Filter Benefits and Limitations}

Learned Bloom filters can easily handle insertions into ${\cal K}$ by
adding the key, if is does not already yield a (false) positive,
to the backup filter.  Such changes have a larger effect on the false
positive probability than for a standard Bloom filter, since the
backup filter is smaller.  Keys cannot be deleted naturally from a
learned Bloom filter.  A deleted key would simply become a false
positive, which (if needed) could possibly be handled by an additional
structure.

As noted in \cite{TCFLIS}, it may be possible to re-learn a new
function $f$ if the data set changes substantially via insertions and
deletion of keys from ${\cal K}$.  Of course, besides the time needed to
re-learn a new function $f$, this requires storing
the original set somewhere, which may not be necessary for alternative schemes.
Similarly, if the false positive probability proves higher than
desired, one can re-learn a new function $f$; again, doing so will
require access to ${\cal K}$, and maintaining a (larger) set ${\cal U}$
of negative examples.  

\section{Size of the Learned Function}

We now consider how to model the performance of the learned Bloom
filter with the goal of understanding how small the representation
of the function $f$ needs needs to be in order for the learned
Bloom filter to be more effective than a standard Bloom filter. \footnote{We
thank Alex Beutel for pointing out that our analysis in \cite{mypaper2} could
be used in this manner.}

Our model is as follows.  The function $f$ associated with
Definition~\ref{def:lbf} we treat as an {\em oracle} for the keys
${\cal K}$, where $|{\cal K}|=m$, that works as follows. For keys not
in ${\cal K}$ there is an associated false positive probability $F_p$,
and there are $F_n m$ false negatives for keys in ${\cal K}$.  (The
value $F_n$ is like a false negative probability, but given ${\cal K}$
this fraction is determined and known according to the oracle
outcomes.)  We note the oracle representing the function $f$ is meant
to be general, so it could potentially represent other sorts of filter
structures as well.  As we have described in Section~\ref{sec:fpr}, in the
context of a learned Bloom filter the false positive rate is
necessarily tied to the query stream, and is therefore generally an
empirically determined quantity, but we take the value $F_p$ here as a given.
Here we show how to optimize over a single oracle, although in
practice we may possibly choose from oracles with different values
$F_p$ and $F_n$, in which case we can optimize for each pair of values
and choose the best suited to the application.

We assume a total budget of $bm$ bits for the backup filter, and $|f| = \zeta$
bits for the learned function.    
If $|{\cal K}| = m$, the backup Bloom filter only needs to hold $mF_n$ keys, and
hence we take the number of bits per stored key to be $b/F_n$.  
To model the false positive rate of a Bloom filter
that uses $j$ bits per stored key, we assume the false positive rate
falls as $\alpha^j$.  This is the case for a standard Bloom filter
(where $\alpha \approx 0.6185$ when using the optimal number of hash
functions, as described in the survey \cite{BroderMitzenmacher}), as
well as for a static Bloom filter built using a perfect hash function
(where $\alpha = 1/2$, again described in \cite{BroderMitzenmacher}).
The analysis can be modified to handle other functions for false
positives in terms of $j$ in a straightforward manner.  (For example,
for a cuckoo filter \cite{cuckoof}, a good approximation for the false
positive rate is $c \alpha^j$ for suitable constants $c$ and $\alpha$.)

The false positive rate of a learned Bloom filter is 
$F_p + (1-F_p)\alpha^{b/F_n}.$
This is because, for $y \notin {\cal
K}$, $y$ causes a false positive from the learned function $f$
with probability $F_p$, or with remaining probability $(1-F_p)$ it
yields a false positive on the backup Bloom filter with probability
$\alpha^{b/F_n}$.

A comparable Bloom filter using the same number of total bits, namely
$bm + \zeta$ bits, would have a false positive probability of
$\alpha^{b+\zeta/m}$.  Thus we find an improvement using a learned Bloom filter
whenever
\begin{align*}
F_p + (1-F_p)\alpha^{b/F_n} \leq \alpha^{b+\zeta/m},
\end{align*}
which simplifies to
\begin{align*}
\zeta/m \leq \log_\alpha \left (F_p + (1-F_p)\alpha^{b/F_n} \right ) - b,
\end{align*}
where we have expressed the requirement in terms of a bound on $\zeta/m$,
the number of bits per key the function $f$ is allowed.  

This expression is somewhat unwieldy, but it provides some insight
into what sort of compression is required for the learned function
$f$, and how a practitioner can determine what is needed.  First, one
can determine possible thresholds and the corresponding rate of false
positive and false negatives from the learned function.  For example,
the paper \cite{TCFLIS} considers situations where $F_p \approx 0.01$,
and $F_n \approx 0.5$; let us consider $F_p =0.01$ and $F_n = 0.5$ for
clarity.  If we have a target goal of one byte per item, a standard
Bloom filter achieves a false positive probability of approximately
$0.0214$.  If our learned function uses 3 bits per item (or less),
then the learned Bloom filter can use $5m$ bits for the backup Bloom
filter, and achieve a false positive rate of approximately $0.0181$.
The learned Bloom filter will therefore provide over a $10\%$
reduction in false positives with the same or less space.  More generally,
in practice one could determine or estimate different $F_p$ and $F_n$ values for 
different thresholds and different learned functions of various sizes, and
use these equations to determine if better performance can be expected without
in depth experiments.  

Indeed, an interesting question raised by this analysis is how learned
functions scale in terms of typical data sets.  
In extreme situations, such as when the set ${\cal
K}$ being considered is a range of consecutive integers, it can be represented by
just two integers, which does not grow with ${\cal K}$.  If, in
practice, as data sets grow larger the amount of information needed for 
a learned function $f$ to approximate key sets ${\cal K}$ grows sublinearly
with $|{\cal K}|$, learned Bloom filters may prove very effective.  

\section{Sandwiched Learned Bloom Filters}

\subsection{The Sandwich Structure}

\begin{figure*}[t]
        \centering
        \includegraphics[width=0.7\textwidth]{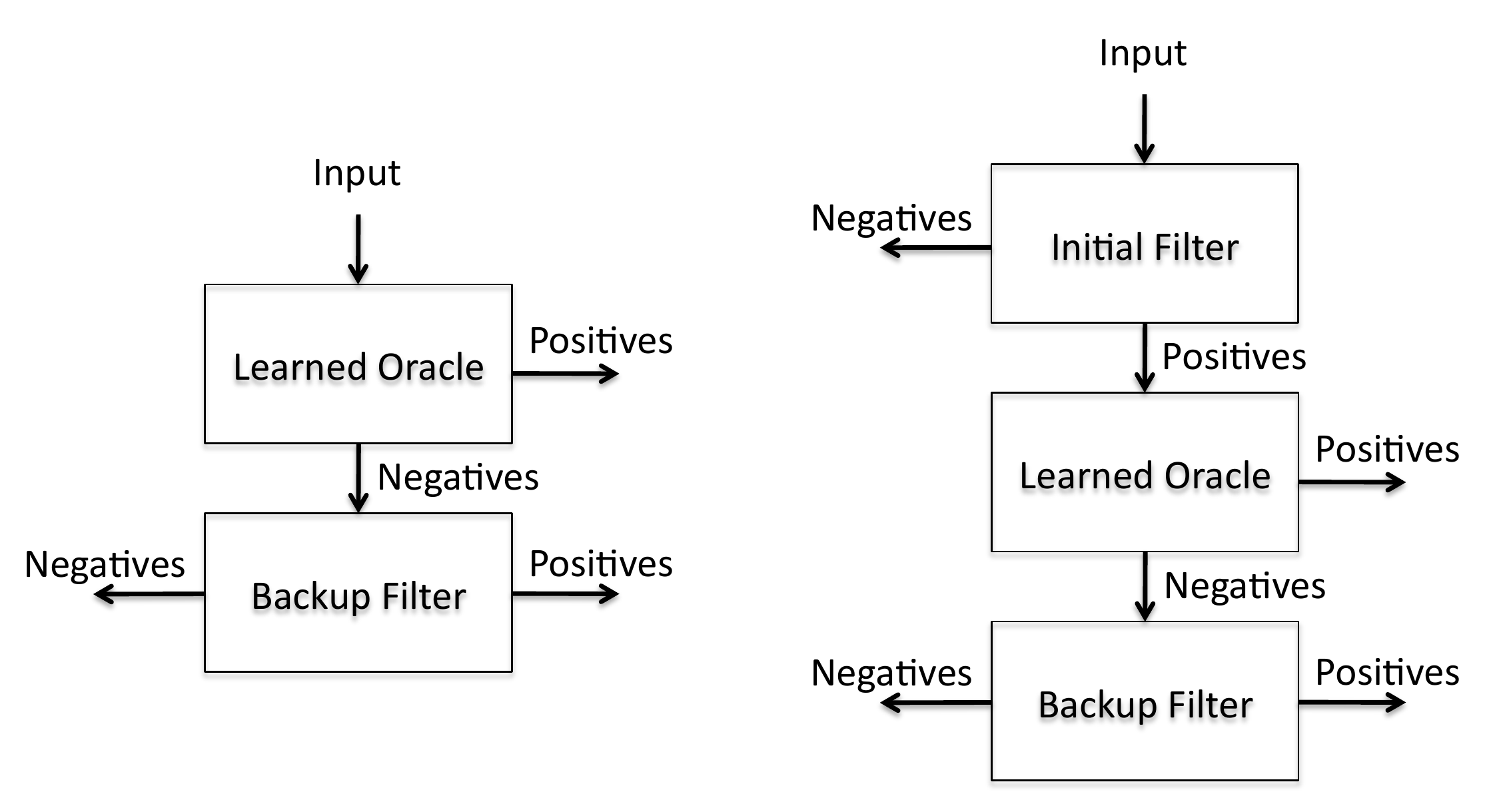}
                \caption{The left side shows the original learned Bloom filter.  The right side shows the sandwiched learned Bloom filter.}
                \label{fig:diagram}
\end{figure*}

Given the formalization of the learned Bloom filter, it seems natural to ask
whether this structure can be improved.  Here we show
that a better structure is to use a Bloom filter before
using the function $f$, in order to remove most queries for keys not
in ${\cal K}$.  We emphasize that this {\em initial Bloom filter} does
not declare that an input $y$ is in ${\cal K}$, but passes forward all
matching keys to the learned function $f$, and it returns $y \notin
{\cal K}$ when the Bloom filter shows the key is not in ${\cal
K}$.  Then, as before, we use the function $f$ to attempt to
remove false positives from the initial Bloom filter, and then use the
backup filter to allow back in keys from ${\cal K}$ that were false
negatives for $f$.  Because we have two layers of Bloom filters
surrounding the learned function $f$, we refer to this as a {\em
sandwiched learned Bloom filter}.  The sandwiched learned Bloom filter
is represented pictorially in Figure~\ref{fig:diagram}.

In hindsight, our result that sandwiching improves performance makes
sense.  The purpose of the backup Bloom filter is to remove the false
negatives arising from the learned function.  If we can arrange to
remove more false positives up front, then the backup Bloom filter can
be quite porous, allowing most everything that reaches it through, and
therefore can be quite small.  Indeed, surprisingly, our analysis shows that the
backup filter should not grow beyond a fixed size.

\subsection{Analyzing Sandwiched Learned Bloom Filters}

We model the sandwiched learned Bloom filter as follows.  As before, the 
learned function $f$ in the middle of the sandwich 
we treat as an {\em oracle} for the keys
${\cal K}$, where $|{\cal K}|=m$. Also as before, for keys not in ${\cal K}$ there is
an associated false positive probability $F_p$, and there are $F_n m$
false negatives for keys in ${\cal K}$.  

We here assume a total budget of $bm$ bits to be divided between an initial
Bloom filter of $b_1m$ bits and a backup Bloom filter of $b_2m$ bits.
As before, we model the false positive rate of a Bloom filter that uses $j$ bits
per stored key as $\alpha^j$ for simplicity.
The backup Bloom filter only needs to hold $mF_n$ keys, and hence we take the number
of bits per stored key to be $b_2/F_n$.  If we find the best
value of $b_2$ is $b$, then no initial Bloom filter is needed, but otherwise,
an initial Bloom filter is helpful.  

The false positive rate of a sandwiched learned Bloom filter is then 
$\alpha^{b_1}(F_p + (1-F_p)\alpha^{b_2/F_n}).$
To see this, note that for $y \notin {\cal K}$, $y$ first has to 
pass through the initial Bloom filter, which occurs with probability 
$\alpha^{b_1}$.  Then $y$ either causes a false positive from the
learned function $f$ with probability $F_p$, or with remaining
probability $(1-F_p)$ it yields a false positive on the backup
Bloom filter, with probability $\alpha^{b_2/F_n}$.

As $\alpha, F_p, F_n$ and $b$ are all constants for the purpose of this analysis,
we may optimize for $b_1$ in the equivalent expression
\begin{align*}
F_p \alpha^{b_1} + (1-F_p)\alpha^{b/F_n}\alpha^{b_1(1-1/F_n)}.
\end{align*}
The derivative with respect to $b_1$ is 
\begin{align*}
F_p (\ln \alpha) \alpha^{b_1} + (1-F_p)\left (1- \frac{1}{F_n} \right)\alpha^{b/F_n}(\ln \alpha)\alpha^{b_1(1-1/F_n)}.
\end{align*}
This equals 0 when 
\begin{eqnarray}
\label{eqn:0deriv}
\frac{F_p}{(1-F_p)\left (\frac{1}{F_n} -1 \right)} & = & \alpha^{(b-b_1)/F_n} = \alpha^{b_2/F_n}.
\end{eqnarray}
This further yields that the false positive rate is minimized when $b_2 = b_2^*$, where
\begin{eqnarray}
\label{eqn:optb}
b_2^* &= & F_n \log_\alpha \frac{F_p}{(1-F_p)\left (\frac{1}{F_n} -1 \right)}.
\end{eqnarray}
This result may be somewhat surprising, as here we see that the optimal
value $b_2^*$ is a constant, independent of $b$.  That is, the number of bits used for
the backup filter is not a constant fraction of the total budgeted
number of bits $bm$, but a fixed number of bits; if the number of
budgeted bits increases, one should simply increase the size of the
initial Bloom filter as long as the backup filter is appropriately
sized.

In hindsight, returning to the expression for the false positive rate
$\alpha^{b_1}(F_p + (1-F_p)\alpha^{b_2/F_n})$ provides useful intuition.
If we think of sequentially
distributing the $bm$ bits among the two Bloom filters, the expression
shows that bits assigned to the initial filter (the $b_1$ bits) reduce
false positives arising from the learned function (the $F_p$ term) 
as well as false positives arising subsequent to the learned function
(the $(1-F_p)$ term), while the backup filter only reduces false
positives arising subsequent to the learned function.
Initially we would provide bits to the backup filter to reduce the $(1-F_p)$ rate
of false positives subsequent to the learned function.  Indeed, 
bits in the backup filter drive down this
$(1-F_p)$ term rapidly, because the backup filter
holds fewer keys from the original set, leading to the $b_2/F_n$ (instead
of just a $b_2$) in the exponent in the expression $\alpha^{b_2/F_n}$.
Once the false positives coming through 
the backup Bloom filter reaches an appropriate level, which, by
plugging in the determined optimal value for $b_2$, we find is $F_p /\left(\frac{1}{F_n} -1 \right)$, then the tradeoff changes.  At that point
the gains from reducing the false positives by increasing the bits for the backup Bloom filter
become smaller than the gains obtained by increasing the bits for the initial Bloom filter.

Again, we can look at situations discussed in \cite{TCFLIS} for some insight.
Suppose we have a learned function $f$ where $F_n =
0.5$ and $F_p = 0.01$.  We consider $\alpha = 0.6185$
(which corresponds to a standard Bloom filter).  We do not consider
the size of $f$ in the calculation below.  Then the optimal value for $b_2$ is 
\begin{align*}
b_2^* = (\log_\alpha 1/99) /2 \approx 6.
\end{align*}

Depending on our Bloom
filter budget parameter $b$, we obtain different levels of performance
improvement by using the initial Bloom filter.  At $b = 8$ bits per
key, the false positive rate drops from approximately $0.010045$ to
$0.005012$, over a factor of 2.  At $b= 10$ bits per key,
the false positive rate drops from approximately $0.010066$ to
$0.001917$, almost an order of magnitude.

We may also consider the implications for the oracle size.  Again, if we 
let $\zeta$ represent the size of the oracle in bits, then a corresponding
Bloom filter would have a false positive probability of $\alpha^{b+\zeta/m}$.
Hence we have an improvement whenever
\begin{align*}
\alpha^{b_1}(F_p + (1-F_p)\alpha^{b_2/F_n}) \leq \alpha^{b+\zeta/m}.
\end{align*}
For $b$ sufficiently large that $b_1 > 0$, we can calculate the false positive
probability of the optimized sandwiched Bloom filter.  Let $b_2^*$ be the
optimal value for $b_2$ from equation~\ref{eqn:optb} and $b_1^*$ be the corresponding
value for $b_1$. First using the relationship from equation~\ref{eqn:0deriv},
we have a gain whenever
\begin{align*}
\alpha^{b_1^*}\frac{F_p}{1-F_n} \leq \alpha^{b+\zeta/m}.
\end{align*}
Using $b_1^* = b - b_2^*$ and equation~\ref{eqn:optb} gives
\begin{align*}
\zeta/m \leq \log_\alpha \frac{F_p}{1-F_n} - F_n \log_\alpha \frac{F_p}{(1-F_p)\left (\frac{1}{F_n} -1 \right)}.
\end{align*}

Again, this expression is somewhat unwieldy, but one useful difference from the analysis
of the original learned Bloom filter is that we see the improvement does not depend on the
exact value of $b$ (as long $b$ is large enough so that $b_1 > 0$, and we use the optimal
value for $b_2$).  
For $F_p =0.01$, $F_n = 0.5$, and $\alpha=0.6185$, we find a gain whenever
$\zeta/m$ falls below approximately $3.36.$

A possible further advantage of the sandwich approach is that it makes
learned Bloom filters more robust. As discussed previously, if
the queries given to a learned Bloom filter do not come from the same
distribution as the queries from the test set used to estimate the
learned Bloom filter's false positive probability, the actual false
positive probability may be substantially larger than expected.  The
use of an initial Bloom filter mitigates this problem, as this issue then only
affects the smaller number of keys that pass the initial Bloom filter.

We note that a potential disadvantage of the sandwich approach may be
that it is more computationally complex than a learned Bloom filter
without sandwiching, requiring possibly more hashing and memory
accesses for the initial Bloom filter.  The overall efficiency would
be implementation dependent, but this remains a possible issue for
further research.  

\eat{
In any case, we suggest that given that the sandwich learned Bloom
filter is a relatively simple modification if one chooses to use a
learned Bloom filter, we believe that the sandwiching method will
allow greater application of the learned Bloom filter methodology.
}

\section{Learned Bloomier Filters}

In the supplemental material, we consider {\em learned Bloomier filters}.
Bloomier filters are a variation of the Bloom filter idea where each key in the set ${\cal K}$ has an associated value.  The Bloomier filter returns the value for every key of ${\cal K}$, and is supposed to return a null value for keys not in ${\cal K}$, but in this context there can be false positives where the return for a key outside of ${\cal K}$ is a non-null value with some probability.  We derive related formulae for the performance of learned Bloomier filters.

\section{Conclusion}

We have focused on providing a more formal analysis of the proposed
learned Bloom filter.  As part of this, we have attempted to clarify a
particular issue in the Bloom filter setting, namely the dependence of
what is referred to as the false positive rate in \cite{TCFLIS} on the
query set, and how it might affect the applications this approach is
suited for.  We have also found that our modeling laeds to a natural
and interesting optimization, based on sandwiching, and allows for
generalizations to related structures, such as Bloomier filters.  Our
discussion is meant to encourage users to take care to realize all of
the implications of the learned Bloom filter approach before adopting
it.  However, for sets that can be accurately predicted by small learned
functions, the learned Bloom filter may provide a novel means of obtaining 
significant performance improvements over standard Bloom filter variants.
 
\newpage

\section*{Acknowledgments}

The author thanks Suresh Venkatasubramanian for suggesting a closer
look at \cite{TCFLIS}, and thanks the authors of \cite{TCFLIS} for
helpful discussions involving their work.  This work was supported in
part by NSF grants CCF-1563710, CCF-1535795, CCF-1320231, and
CNS-1228598.  Part of this work was done while visiting Microsoft
Research New England.

\newpage

\section{Supplemental:  Learned Bloomier Filters Derivation}

Bloomier filters are a variation of the Bloom filter idea where each key in the set ${\cal K}$ has an associated value, which for convenience we will assume is a $u$-bit value.  The Bloomier filter returns the value for every key of ${\cal K}$, as is supposed to return a null value for keys not in ${\cal K}$, but in this context there can be false positives where the return for a key outside of ${\cal K}$ is a non-null value with some probability.  For our purposes, this description will suffice (although we present a few more details below in presenting our model), but more information on Bloomier filters and their constructions can be found in \cite{charles2008bloomier,chazelle2004bloomier}.  

Here we imagine that we can derive a learned function $f$ that will return a value given an input, with the goal being that the function will return the appropriate $u$-bit value for a key in ${\cal K}$ and the null value otherwise.  In this setting we refer to a false positive as a key outside of ${\cal K}$ that obtains a non-null value, and a false negative as a key in ${\cal K}$ that obtains an incorrect value, where that value may either be the null value or the wrong value for that key.  

\begin{figure}[t]
        \centering
        \includegraphics[width=0.4\textwidth]{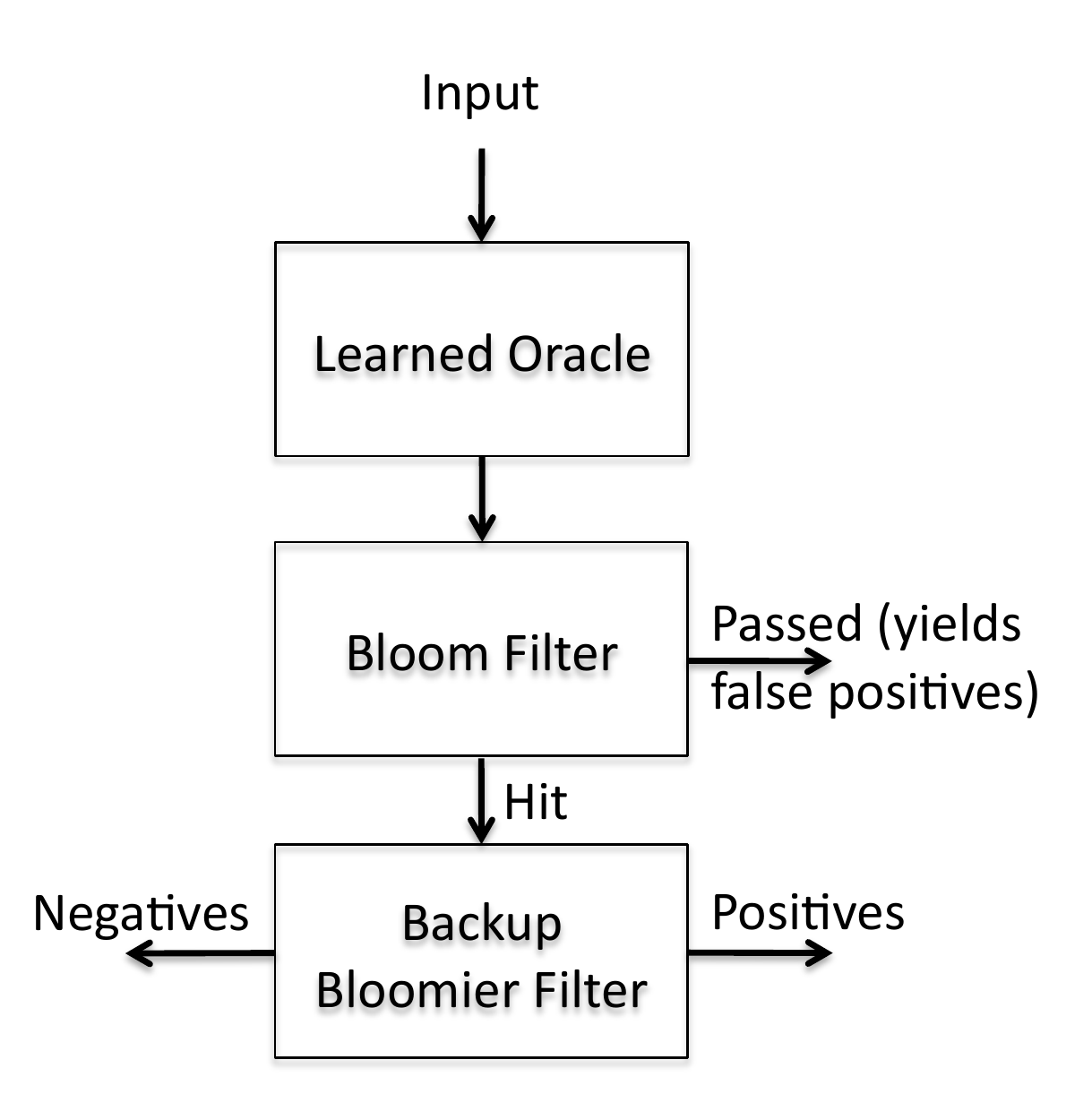}
                \caption{A learned Bloomier filter design.  Keys obtain a value, which may be null, from the learned oracle.  If a key does {\em not} hit on a Bloom filter, the key returns with the value from the oracle;  in this way, false positives from the oracle may result.  The Bloom filter stores false negatives from the learned oracle, and passes them to the backup Bloomier filter to obtain the correct value.  False positive hits at the Bloom filter both require the backup Bloomier filter to hold values for additional keys from ${\cal K}$, and may yield false positives for keys outside of ${\cal K}$ at the backup Bloomier filter.}
                \label{fig:two}
\end{figure}

Notice in this setting that, because keys in ${\cal K}$ may obtain an incorrect value that is not merely null, the system to correct for false negatives must be slightly more complicated.  We propose an approach in Figure~\ref{fig:two}.  The input is first passed to the learned oracle, which provides a predicted value.  To handle false negatives, we provide a two-stage scheme.  First, we use a Bloom filter to hold any keys that lead to false negatives.  If the Bloom filter returns a key is a positive, which we refer to as a hit on the Bloom filter to avoid ambiguity, it is assumed that that key is a false negative from the oracle and a backup Bloomier filter is used to determine its value.  If the Bloom filter returns a key is a negative, it is assumed the learned oracle provided the correct value (whether null or non-null) for that key, and that value is returned.  We can see that for every key in ${\cal K}$ the correct value is returned, so the question is what is the false positive rate for this chain.  

Our model is as follows.  We again treat the function $f$ as an oracle
for the keys in ${\cal K}$, where $|{\cal K}|=m$, and the size of the
oracle is $\zeta$.  For keys not in ${\cal K}$ there is an associated
false positive probability $F_p$, and there are $F_n m$ false
negatives for keys in ${\cal K}$.  The Bloom filter will consist of $b
F_n m$ bits and have its own false positive probability of $\alpha^b$.
(Of course, instead of a standard Bloom filter we could use a learned
Bloom filter in its place, but that is harder to model.)

To model a Bloomier filter, we use the following approach: a Bloomier
filter for $z$ keys uses space $cz(u+r)$, where $u$ is the number
of bits in the return value, $c$ is constant that is determined by the
Bloomier filter design, and $r$ is a parameter chosen, with the result
that the false positive probability for a key outside of ${\cal K}$ is
$2^{-r}$.  This setup, for example, matches the construction
of \cite{chazelle2004bloomier}.  One can think of it as having $cz$
cells of $u+r$ bits.  The simple construction of \cite{chazelle2004bloomier}
hashes a query key to (typically) three cells, and exclusive-ors their
contents together;  if the result is a valid $u$-bit value (say with $r$ leading
zeroes), this value is returned, and otherwise a null value is returned.  
The table is initially filled so that the right values are returned for the
$z$ keys, and other keys obtain a value uniformly distributed over
$u+r$ bits, leading to the false positive probability of $2^{-r}$.  This requires
$cz$ cells for some $c > 1$ to provide enough ``room'' to set up suitable cell
values initially.  

If we just used a standard Bloomier filter for the keys ${\cal K}$, then
we would use $cm(u+r)$ bits for a false positive probability of $2^{-r}$.  

For our learned Bloom filter construction, we start with the learned
function of size $\zeta$.  The function $f$ yields $mF_n$ false
negatives; these $mF_n$ false negatives can be stored in the Bloom
filter using $bmF_n$ bits and the corredsponding values recovered by
the backup Bloomier filter.  The keys not hit by this Bloom filter,
which we refer to as the keys passed by this Bloom filter, may now
include false positives for our learned Bloomier filter.  A key not in ${\cal
K}$ will yield a false positive here with probability
$F_p(1-\alpha^b)$; that is, the key must have been a false positive
for the learned oracle, but must have not been a hit on the Bloom
filter.  Also, note that some keys from ${\cal K}$ that obtained the
correct value from $f$ may hit the Bloom filter, and therefore will also have
to have their values provided by the backup Bloomier filter.  Of these
$m(1-F_n)$ keys from ${\cal K}$, a fraction $\alpha^b$ are expected to be
false positives in the Bloom filter;  as we have throughout the paper, we
will use the expectation, keeping in mind the true result will be concentrated
around this expectation.  Hence, in total, we need the backup Bloomier filter
for $m' = m(F_n + (1-F_n)\alpha^b)$ keys.  Suppose we use $cm'(u+r')$ bits
for the backup Bloomier filter.  Then our total space is
\begin{align*}
\zeta + bmF_n +cm(F_n + (1-F_n)\alpha^b)(u+r'),
\end{align*}
and our overall false positive probability is 
\begin{align*}
F_p(1-\alpha^b)+\alpha^b 2^{-r'},
\end{align*}
where the first term is from false positives from the oracle than passed Bloom filter,
and the second term is from queries that hit the Bloom filter and give false positives
in the backup Bloomier filter.  

Again, these expressions are somewhat unwieldy because of the number
of parameters.  At a high level, however, these expressions reinforce
helpful intuition.  The cost per element in a Bloomier filter is
rather high, because the value must be stored.  Therefore if the false
negatives as given by $F_n$ can be driven down to reasonable value with a small
enough learned function, there should be space available to pay the $\zeta$ bits
of the learned function as well as the additional Bloom filter.

\end{document}